\documentclass{article}
\usepackage{spconf}
\usepackage{amsmath}
\usepackage{graphicx}
\usepackage{subfigure}
\usepackage{cite}
\usepackage{bbm}
\usepackage{amssymb}
\usepackage{multirow}
\usepackage{amsthm}
\usepackage{bm}
\usepackage{cite}
\usepackage{algorithm}
\usepackage{algorithmic}
\usepackage{url}

\def \X {{\mathbf{X}}}
\def \A {{\mathbf{A}}}
\def \D {{\mathbf{D}}}
\def \L {{\mathbf{L}}}

\def \x {{\mathbf{x}}}

\def \ones {{\mathbf{1}}}

\title{CENTRALITY-CONSTRAINED GRAPH EMBEDDING}

\name{Brian Baingana, Georgios B. Giannakis}
\address{Dept. of ECE, University of Minnesota, Minneapolis, MN 55455, USA}

\begin{document}
\ninept

\maketitle
\begin{abstract}
\indent Visual rendering of graphs is a key task in the mapping of
complex network data. Although most graph drawing algorithms
emphasize aesthetic appeal, certain applications such as travel-time
maps place more importance on visualization of structural network
properties. The present paper advocates a graph embedding approach
with centrality considerations to comply with node hierarchy. The
problem is formulated as one of constrained multi-dimensional
scaling (MDS), and it is solved via block coordinate descent
iterations with successive approximations and guaranteed convergence
to a KKT point. In addition, a regularization term enforcing graph
smoothness is incorporated with the goal of reducing edge crossings.
Experimental results demonstrate that the algorithm converges, and
can be used to efficiently embed large graphs on the order of
thousands of nodes.
\end{abstract}
\noindent
\section{Introduction}
\label{sec:intro}
Graphs offer a valuable means of encoding
relational information between entities of complex systems, arising
in modern communications, transportation and social networks, among
others. Despite the abundance of network analysis techniques,
information visualization is a powerful tool for capturing patterns
that may not be apparent in large-scale systems. However, most
visualization algorithms focus more on aesthetic appeal than the
structural characteristics of the underlying data. Such network
structure is captured through graph-theoretic notions such as node
centrality and network cohesion.

The present paper deals with embedding graphs for visualization
while adhering to the underlying node centrality structure.
Centrality measures capture the relative importance of network nodes
among their peers. Betweenness centrality for instance, describes
the extent to which information is routed through a specific node by
measuring the fraction of all shortest paths traversing this node;
see e.g.,~\cite[p. 89]{netsci}. Other measures include closeness,
eigenvalue, and degree centrality. To incorporate centrality using
any of these metrics, an MDS (so-termed stress~\cite[Chap. 3]{mds}) 
criterion is adopted, under radial constraints that place nodes of higher
centrality closer to the origin of the graph embedding. MDS seeks a
low-dimensional depiction of high-dimensional data in which pairwise
Euclidean distances between embedding coordinates are close (in a
well-defined sense) to the dissimilarities between the original data
points. Closeness criteria (a.k.a. stress costs) are generally
non-convex, and the quest for global optimality is challenging
because ordinary descent methods do not have optimality guarantees,
and are sensitive to initialization. Successive approximation with
global and convex upper bounds is used in \cite[Chap. 8]{mds} to 
minimize the stress cost yielding near-optimal results.

The novel approach exploits the block separability inherent to the
proposed model and adapts the coordinate descent algorithm to
determine the optimal embedding. Edge crossings are minimized by
regularizing the cost with a smoothness promoting term weighted by a
tuning parameter. Smoothness encourages nodes that share an edge to
lie closer to each other in the embedding. As a result, the length
and hence the number of edge crossings in the network visualization
is markedly reduced. In addition, the regularization term offers the
benefit of incorporating the underlying network topology when the
dissimilarities considered are not graph-theoretic e.g., Euclidean
distances between feature vectors associated with each node.
Moreover, numerical tests illustrate that judicious selection of the
tuning parameter results in fewer block coordinate descent
iterations, which in turn yields a visually appealing embedding.

To place the present work in context, a prior approach iteratively
minimizes a weighted stress function with iteration-dependent
weights chosen to incorporate radial constraints \cite{brandes}.
However, it is limited to graph-theoretic dissimilarities, and
offers no convergence guarantees. A heuristic algorithm for network
visualization uses the $k$-core decomposition to hierarchically
place nodes within ``onion-like'' concentric shells \cite{finger}.
Although effective for large-scale networks, it has no optimality
associated with it, and is limited to visualization only in $2$
dimensions. The proposed approach scales well for large networks
under a well-defined optimality criterion with a convergence
guarantee.

\section{Model and Problem Statement}
\label{model}

Consider a network represented by an undirected graph $\mathcal{G} =
(\mathcal{V}, \mathcal{E})$, where $\mathcal{E}$ denotes the set of
edges, and $\mathcal{V}$ the set of vertices with cardinality
$|\mathcal{V}| = N$. Let $\delta_{ij}$ denote the pairwise
dissimilarity (edge weight) between any two nodes $i$ and $j$. Given
the set $\{\delta_{ij}\}$ and the prescribed embedding dimension
$p$, the graph embedding task amounts to finding $p \times 1$
vectors $\left\lbrace \x_i \right\rbrace_{i=1}^N $ so that the
embedding coordinates $\x_i$ and $\x_j$ satisfy $ \| \x_i - \x_j
\|_2 \approx \delta_{ij}$.

With $\delta_{ij} = \delta_{ji}$, it suffices to know $\{ \{
\delta_{ij} \}_{j=1}^N \}_{i=j+1}^N$, or, be possible to determine
them from $\mathcal{G}$. Most visualization schemes assign
$\delta_{ij}$ to the shortest path distance between nodes $i$ and
$j$. In this work, the Euclidean commute-time distance (ECTD) is
adopted because it decreases as the number of shortest paths between
node pairs increases \cite{fouss}. This is more reasonable since
having multiple shortest paths between node pairs endows them with a
higher level of accessibility by e.g., a random walker on the graph.

MDS amounts to solving the following problem:
\begin{equation}
\label{eq1} (\text{P}0)\;\;\;\; \left\lbrace \hat{\x}_i
\right\rbrace_{i=1}^N =  \underset{\x_1, \dots,
\x_N}{\operatorname{arg\,min}} \text{ }\frac{1}{2} \sum_{i =
1}^{N}\sum_{j = 1}^{N}[\left\| \mathbf{x}_i - \mathbf{x}_j
\right\|_2 - \delta_{ij}]^2.
\end{equation}

Turning attention to node centralities $\{ c_i \}_{i=1}^N$, those
can be obtained using a number of algorithms~\cite[Chap. 4]{netsci}.
Centrality structure will be imposed on \eqref{eq1} by constraining
$\x_i$ to have a centrality-dependent radial distance $f(c_i)$,
where $f(.)$ is a monotone decreasing function. The resulting
constrained optimization problem now becomes
\begin{eqnarray}
\label{eq2} \nonumber (\text{P}1)\;\;\;\; \left\lbrace \hat{\x}_i
\right\rbrace_{i=1}^N = \underset{\x_1, \dots,
\x_N}{\operatorname{arg\,min}} &
  \frac{1}{2}
  \sum\limits_{i = 1}^{N}\sum\limits_{j = 1}^{N}
  \left[
   \left\| \mathbf{x}_i - \mathbf{x}_j \right\|_2 - \delta_{ij}
  \right]^2 \\
 \text{s. to} &  \left\| \mathbf{x}_i \right\|_2 = f(c_i),\text{ }i = 1,\dots,N.
\end{eqnarray}

Although $\text{P}0$ is non-convex, standard solvers rely on
gradient descent iterations but have no guarantees of convergence to
the global optima \cite{buja}. Lack of convexity is exacerbated in
$\text{P}1$ by the non-convex constraint set rendering its solution
even more challenging than that of $\text{P}0$. However, considering
a single embedding vector $\x_i$, and fixing the rest $\{\x_j\}_{j
\ne i}$, the constraint set simplifies to $\| \x_i \|_2 = f(c_i)$,
for which an appropriate relaxation can be sought. Key to the
algorithm proposed next lies in this inherent decoupling of the
centrality constraints.

\section{BCD with successive approximations}
By exploiting the separable nature of the cost as well as the norm
constraints in \eqref{eq2}, block coordinate descent (BCD) will be
adopted in this section to arrive at a solution approaching the
global optimum. To this end, the centering constraint $\sum_{i=1}^N
\x_i = {\bf 0}$, typically invoked to fix the inherent translation
ambiguity, will be dropped first so that the problem remains
decoupled across nodes. The effect of this relaxation can be
compensated for by computing the centroid of the solution of
\eqref{eq2}, and subtracting it from each coordinate. The $N$
equality norm constraints are also relaxed to $\| \x_i \|_2 \le
f(c_i)$. Although the entire constraint set is non-convex, each
relaxed constraint is a convex and closed Euclidean ball with
respect to each node in the network.

Let $\x_i^r$ denote the minimizer of the optimization problem over
block $i$, when the remaining blocks $\left\lbrace
\x_j\right\rbrace_{j \ne i}$ are fixed during the BCD iteration $r$.
By fixing the blocks $\left\lbrace \x_j \right\rbrace_{j \ne i}$ to
their values from the most recent iterations, the sought embedding
is obtained as
\begin{equation}
\label{eq3}
\left\lbrace \hat{\x}_i \right\rbrace_{i=1}^N =
\underset{\left\lbrace \x_i : \left\| \mathbf{x}_i \right\|_2
\le f(c_i) \right\rbrace}{\operatorname{arg\,min}}
  \frac{1}{2}
  \sum\limits_{i = 1}^{N}\sum\limits_{j = 1}^{N}
  \left[
   \left\| \mathbf{x}_i - \mathbf{x}_j \right\|_2 - \delta_{ij}
  \right]^2
\end{equation}
or equivalently as
\begin{eqnarray}
\label{eq4}
\nonumber
\underset{\left\lbrace \x_i : \left\| \mathbf{x}_i \right\|_2 \le f(c_i) \right\rbrace}{\arg\min}
\frac{(N-1)}{2}\| \x_i \|_2^2 - \x_i^T( \sum\limits_{j < i} \x_j^r + \sum\limits_{j > i} \x_j^{r-1}) \\
- \sum\limits_{j < i} \delta_{ij} \| \x_i - \x_j^r \|_2 - \sum\limits_{j > i} \delta_{ij} \| \x_i - \x_j^{r-1} \|_2
\end{eqnarray}
where $ \sum\limits_{j < i} (.) := \sum\limits_{j = 1}^{i-1} (.)$
and $ \sum\limits_{j > i} (.) := \sum\limits_{j = i+1}^{N} (.)$.
With the last two sums in \eqref{eq4} being non-convex and
non-smooth, convergence of the BCD algorithm cannot be guaranteed
\cite[p. 272]{bertsekas}. Moreover, it is desired to have each per-iteration
subproblem solvable to global optimality, in closed form and at a
minimum computational cost. The proposed approach seeks a global
upper bound of the objective with the desirable properties of
smoothness and convexity. To this end, consider the function $\Psi(
\x_i) := \psi_1(\x_i) - \psi_2 (\x_i)$, where
\begin{equation}
\label{eq4a} \psi_1(\x_i) := \frac{(N-1)}{2}\| \x_i \|_2^2 - \x_i^T(
\sum\limits_{j < i} \x_j^r + \sum\limits_{j > i} \x_j^{r-1})
\end{equation}
and
\begin{equation}
\label{eq4b} \psi_2 (\x_i) :=  \sum\limits_{j < i} \delta_{ij} \|
\x_i - \x_j^r \|_2 + \sum\limits_{j > i} \delta_{ij} \| \x_i -
\x_j^{r-1} \|_2.
\end{equation}

Note that $ \psi_1(\x_i) $ is a convex quadratic function, and that
$\psi_2(\x_i) $ is convex (with respect to $\x_i$) but
non-differentiable. The first-order approximation of \eqref{eq4b} at
any point in its domain is a global under-estimate of $\psi_2(\x_i)
$. Despite the non-smoothness at some points, such a lower bound can
always be established using its subdifferential. As a consequence of
the convexity of $\psi_2(\x_i) $, it holds that~\cite[p.~731]{bertsekas}
\begin{equation}
\label{eq5}
\psi_2(\x)  \ge  \psi_2(\x_0) +
\mathbf{g}^T(\x_0)(\x - \x_0),  \forall \x \in \text{dom}(\psi_2)
\end{equation}
where $\mathbf{g}(\x) \in \partial \psi_2(\x)$ is a subgradient
within the subdifferential set, $\partial \psi_2(\x)$ of $
\psi_2(\x)$. The subdifferential of $\| \x_i - \x_j \|_2$ with
respect to $\x_i$ is given by
\begin{equation}
\label{eq6}
\partial_{\x_i} \| \x_i - \x_j \|_2 =
\begin{cases}
\frac{\x_i - \x_j}{\| \x_i - \x_j \|_2}, & \text{ if } \x_i \ne \x_j \\
\mathbf{y} \in \mathbb{R}^p : \text{ } \| \mathbf{y} \|_2 \le 1, & \text{ otherwise }
\end{cases}
\end{equation}
which implies that
\begin{equation}
\label{eq7}
\partial_{\x_i} \psi_2(\x_i) = \sum\limits_{j = 1}^N \delta_{ij} \partial_{\x_i} \| \x_i - \x_j \|_2.
\end{equation}
Using \eqref{eq5}, it is possible to lower bound \eqref{eq4b} by
\begin{eqnarray}
\label{eq8}
\nonumber
\psi_2'(\x_i, \x_0) = \sum\limits_{j < i} \delta_{ij} \left[ \| \x_0 - \x_j^r \|_2 +
 (\mathbf{g}_j^{r})^T(\x_0)(\x_i - \x_0) \right]  \\
  + \sum\limits_{j > i} \delta_{ij} \left[  \| \x_0 - \x_j^{r-1} \|_2
+ (\mathbf{g}_j^{r-1})^T(\x_0)(\x_i - \x_0)\right].
\end{eqnarray}
Consider now $\Phi(\x_i, \x_0) := \psi_1(\x_i) - \psi_2'(\x_i,
\x_0)$, and note that $\Phi(\x_i, \x_0)$ is convex and upper bounds
globally the cost in \eqref{eq4}. The proposed BCD algorithm
involves successive approximations using \eqref{eq8}, and yields the
following QCQP for each block
\begin{equation}
\label{eq9} (\text{P}2) \quad\quad \underset{\left\lbrace \x_i :
\left\| \mathbf{x}_i \right\|_2 \le f(c_i)
\right\rbrace}{\operatorname{arg\,min}}  \Phi(\x_i, \x_0).
\end{equation}
For convergence, $\x_0$ must be selected to satisfy the following
conditions \cite{meis}:
\begin{subequations}
\begin{eqnarray}
\label{eq10}
\Phi(\x_0, \x_0) & = & \Psi(\x_0), \quad \forall \x_0 \in \mathcal{C},  \forall i  \\
\label{eq10b} \Phi(\x_i, \x_0) & \ge & \Psi(\x_i), \quad \| \x_i
\|_2 \le f(c_i),  \forall i
\end{eqnarray}
\end{subequations}
where $\mathcal{C} := \bigcup_{i = 1}^{N} \left\lbrace \x_i : \|
\x_i \|_2 \le f(c_i)\right\rbrace $. In addition, $\Phi(\x_i, \x_0)$
must be continuous in $(\x_i, \x_0)$. Upon selecting $\x_0 =
\x_i^{r-1}$, the iterate $\x_i^{r-1}$ satisfies \eqref{eq10} and
\eqref{eq10b}. Taking successive approximations around $\x_i^{r-1}$
in $\text{P}2$, ensures the uniqueness of
\begin{eqnarray}
\label{eq11}
\nonumber
\x_i^r &=\underset{\left\lbrace \x_i : \left\| \mathbf{x}_i \right\|_2
\le f(c_i)  \right\rbrace}{\operatorname{arg\,min}}  \frac{(N-1)}{2} \x_i^T\x_i \\
\nonumber
&- \x_i^T[ \sum\limits_{j < i}(\x_j^r + \delta_{ij}\mathbf{g}_j^r(\x_i^{r-1}) ) \\
&+ \sum\limits_{j > i}(\x_j^{r-1} + \delta_{ij}\mathbf{g}_j^{r-1}(\x_i^{r-1}) ) ]
\end{eqnarray}
Solving \eqref{eq11} amounts to obtaining the solution of the
unconstrained QP, $(\x_i^*)^r$, and projecting it onto $\left\lbrace
\x_i : \left\| \mathbf{x}_i \right\|_2 \le f(c_i) \right\rbrace$;
that is,
\begin{equation}
\label{eq12}
\x_i^r =
\begin{cases}
\frac{(\x_i^*)^r}{\| (\x_i^*)^r \|_2} f(c_i),\;\; \text{if}\;\;\; \| (\x_i^*)^r \|_2 > f(c_i) \\
(\x_i^*)^r , \text{ otherwise }
\end{cases}
\end{equation}
where
\setlength{\abovedisplayskip}{3pt}
\setlength{\belowdisplayskip}{3pt}
\begin{eqnarray}
\label{eq13}
\nonumber
(\x_i^*)^r & = \frac{1}{N-1}[ \sum\limits_{j < i}(\x_j^r + \delta_{ij}\mathbf{g}_j^r(\x_i^{r-1}) ) \\
&+ \sum\limits_{j > i}(\x_j^{r-1} + \delta_{ij}\mathbf{g}_j^{r-1}(\x_i^{r-1}) ) ].
\end{eqnarray}
It is desirable but not necessary that the algorithm converges
because depending on the application, reasonable network
visualizations can be found with fewer iterations. In fact,
successive approximations merely provide a more refined graph
embedding that maybe more aesthetically appealing.

Although the proposed algorithm is guaranteed to converge, the
solution is only unique up to a rotation and a translation (cf.
MDS). In order to eliminate the translational ambiguity, the
embedding can be centered at the origin. Assuming that the optimal
blocks determined within outer iteration $r$ are reassembled into
the embedding matrix $\X^r := \left[ (\x_1^r)^T, \dots, (\x_N^r)^T \right]^T$,
the final step involves subtracting the mean from each coordinate
using the centering operator as follows, $\X = (\mathbf{I} -
N^{-1}\ones \ones^T)\X^r$, where $\mathbf{I}$ denotes the $N \times
N$ identity matrix, and $\ones$ is the $N \times 1$ vector of all
ones.

The novel graph embedding scheme is summarized as Algorithm
\ref{alg1} with matrix $\mathbf{\Delta}$ having $(i,j)$th entry the
dissimilarity $\delta_{ij}$.
\begin{algorithm}
    \caption{BCD algorithm with successive approximations}
\label{alg1}
\begin{algorithmic}
   \STATE {\bfseries Input:}  $\left\lbrace c_i \right\rbrace_{i=1}^N$,  $\mathbf{\Delta}$, $\epsilon$
   \STATE Initialize $\X^0$, $r=0$
   \REPEAT
   \STATE $r = r + 1$
   \FOR{$i=1 \dots N$}
   \STATE Compute $\x_i^r$ according to \eqref{eq12} and \eqref{eq13}
   \STATE $\X^r(i,:) = (\x_i^r)^T$
   \ENDFOR
   \UNTIL{$\| \X^{r} - \X^{r-1} \|_F \le \epsilon$}
   \STATE $\X = (\mathbf{I} - \frac{1}{N}\ones \ones^T)\X^r$
\end{algorithmic}
\end{algorithm}

\section{Enforcing graph smoothness}
\label{section:smoothness} In this section, the MDS stress in
\eqref{eq3} is regularized through an additional constraint that
encourages smoothness over the graph. Intuitively, despite the
requirement that the node placement in low-dimensional Euclidean
space respects inherent network structure, through preserving e.g.,
node centralities, neighboring nodes in a graph-theoretic sense
(meaning nodes that share an edge) are expected to be close in
Euclidean distance within the embedding. Such a requirement can be
captured by incorporating a constraint that discourages large
distances between neighboring nodes. In essence, this constraint
enforces smoothness over the graph embedding.

A popular choice of a smoothness-promoting function is $ h(\X) :=
\text{Tr}(\X^T\L\X)$, where $\text{Tr}(.)$ denotes the trace
operator, and $\L := \D - \A$ is the graph Laplacian with $\D$ a
diagonal matrix whose $(i,i)$th entry is the degree of node $i$, and
$\A$ the adjacency matrix. It can be shown that $h(\X) = (1/2)
\sum\limits_{i = 1}^{N} \sum\limits_{i = 1}^{N} a_{ij} \| \x_i -
\x_j \|_2^2$, where $a_{ij}$ is the $(i,j)$th entry of $\mathbf{A}$.
Motivated by penalty methods in optimization, the cost in
\eqref{eq2} will be augmented as follows
\begin{eqnarray}
\label{eq15} \nonumber (\text{P}3)\;\;\; \quad
\underset{\mathbf{\x_1, \dots, \x_N}}{\operatorname{arg\,min}} &
  \frac{1}{2}
  \sum\limits_{i = 1}^{N}\sum\limits_{j = 1}^{N}
  \left[
   \left\| \mathbf{x}_i - \mathbf{x}_j \right\|_2 - \delta_{ij}
  \right]^2  \\
  \nonumber
  & +  \frac{\lambda}{2} \sum\limits_{i=1}^{N} \sum\limits_{j=1}^{N} a_{ij}\| \x_i - \x_j \|_2^2
  \\
\text{s. to} & \left\| \mathbf{x}_i \right\|_2 = f(c_i), i = 1,\dots,N
\end{eqnarray}
where the scalar $\lambda \ge 0$ controls the degree of smoothness.
The penalty term has a separable structure and is convex with
respect to $\x_i$. Consequently, $\text{P}3$ lies within the
framework of successive approximations required to solve each
per-iteration subproblem. Following the same relaxations and
invoking the successive upper bound approximations described
earlier, yields the following QCQP
\begin{eqnarray}
\label{eq16}
\nonumber
\x_i^r &=\underset{\left\lbrace \x_i : \left\| \mathbf{x}_i \right\|_2
\le f(c_i)  \right\rbrace}{\operatorname{arg\,min}}
(N + \lambda d_{ii} - 1) \x_i^T\x_i \\
\nonumber
&- \x_i^T[ \sum\limits_{j < i}((1+\lambda a_{ij})\x_j^r + \delta_{ij}\mathbf{g}_j^r(\x_i^{r-1}) ) \\
&+ \sum\limits_{j > i}((1+\lambda a_{ij})\x_j^{r-1} + \delta_{ij}\mathbf{g}_j^{r-1}(\x_i^{r-1}) ) ]
\end{eqnarray}
with $d_{ii} := \sum\limits_{j=1}^{N}a_{ij}$
denoting the degree of node $i$. \\
\indent The solution of \eqref{eq16} can be expressed as [cf.
\eqref{eq12}]
\begin{eqnarray}
\label{eq17}
\nonumber
(\x_i^*)^r &=\frac{1}{N + \lambda d_{ii} -1}[ \sum\limits_{j < i}((1+\lambda a_{ij})\x_j^r + \delta_{ij}\mathbf{g}_j^r(\x_i^{r-1}) ) \\
&+ \sum\limits_{j > i}((1+\lambda a_{ij})\x_j^{r-1} + \delta_{ij}\mathbf{g}_j^{r-1}(\x_i^{r-1}))].
\end{eqnarray}
With $\lambda$ given, Algorithm \ref{alg2} summarizes the steps to
determine the constrained embedding with a smoothness penalty.
\begin{algorithm}
    \caption{Incorporating smoothness in Algorithm 1}
\label{alg2}
\begin{algorithmic}
   \STATE {\bfseries Input:}  $\A$, $\left\lbrace c_i \right\rbrace_{i=1}^N$,  $\mathbf{\Delta}$, $\epsilon$, $\lambda$
   \STATE Initialize $\X^0$, $r=0$
   \REPEAT
   \STATE $r = r + 1$
   \FOR{$i=1 \dots N$}
   \STATE Compute $\x_i^r$ according to \eqref{eq12} and \eqref{eq17}
   \STATE $\X^r(i,:) = (\x_i^r)^T$
   \ENDFOR
   \UNTIL{$\| \X^{r} - \X^{r-1} \|_F \le \epsilon$}
   \STATE $\X = (\mathbf{I} - \frac{1}{N}\ones \ones^T)\X^r$
\end{algorithmic}
\end{algorithm}
\vspace{-0.3cm}
\section{Numerical Experiments}
\label{experiments}
\subsection{Visualizing the London Tube}
\label{ltube} 
\begin{figure}[t]
\centering
\includegraphics[scale=0.38]{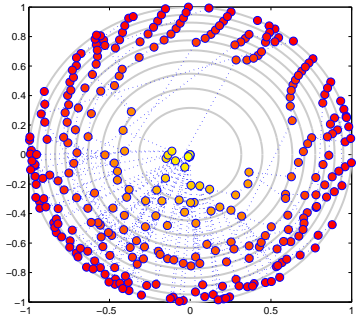}
\vspace*{-0.5cm}
\caption{Centrality-constrained embedding of the London tube}
\vspace*{-0.4cm}
\label{fig:london_tube_no_smoothness}
\end{figure}

\begin{figure}[t]
\centering
\includegraphics[scale=0.38]{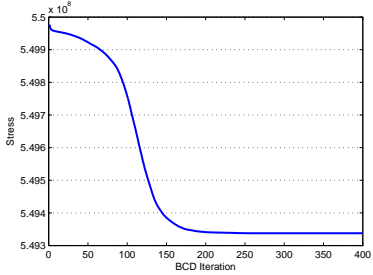}
\vspace*{-0.5cm}
\caption{MDS stress iterations}
\label{fig:london_tube_Stress_vs_iter}
\vspace*{-0.5cm}
\end{figure}

In the first experiment, an undirected graph of $307$
nodes representing the London tube, an underground train transit
network,\footnote{https://wikis.bris.ac.uk/display/ipshe/London+Tube}
is considered. The nodes represent stations whereas the edges
represent the routes connecting them. The objective is to generate
an embedding in which stations traversed by most routes are placed
closer to the center, thus highlighting their relative significance
in metro transit. Such information is best captured by the
\emph{betweenness centrality}, which is defined as  
$ c_i := \sum_{j \ne k \ne i \in \mathcal{V}}
\sigma_{j,k}^{i}/(\sum_{i \in \mathcal{V}}\sigma_{j,k}^{i})$, 
where $\sigma_{j,k}^{i}$ is the number of shortest paths between nodes 
$j$ and $k$ through node $i$~\cite{freeman}. The centrality values
were transformed as follows:
\begin{equation}
\label{eqq2} f\left( c_i \right) = \frac{\text{diam} \left(
\mathcal{G} \right) }{2} \left( 1 - \frac{c_i - \underset{i \in
\mathcal{V}}{\operatorname{min}} \text{ } c_i}{ \underset{i \in
\mathcal{V}}{\operatorname{max}}\text{ }c_i - \underset{i \in
\mathcal{V}}{\operatorname{max}}\text{ }c_i} \right)
\end{equation}
with $\text{diam}( \mathcal{G})$ denoting the diameter of
$\mathcal{G}$. Simulations were run for several values of $\lambda$
starting with $\lambda = 0$, and the resultant two-dimensional
($p=2$) embeddings were plotted. Figure
\ref{fig:london_tube_no_smoothness} depicts the optimal embedding
obtained without a smoothness penalty. The color grading reflects
the centrality levels of the nodes from highest (yellow) to lowest
(red). Algorithm 1 converged after approximately $150$ outer
iterations
as shown in Figure \ref{fig:london_tube_Stress_vs_iter}.

\begin{figure}[htb]
\begin{minipage}[b]{.48\linewidth}
  \centering
  \centerline{\includegraphics[width=3.3cm]{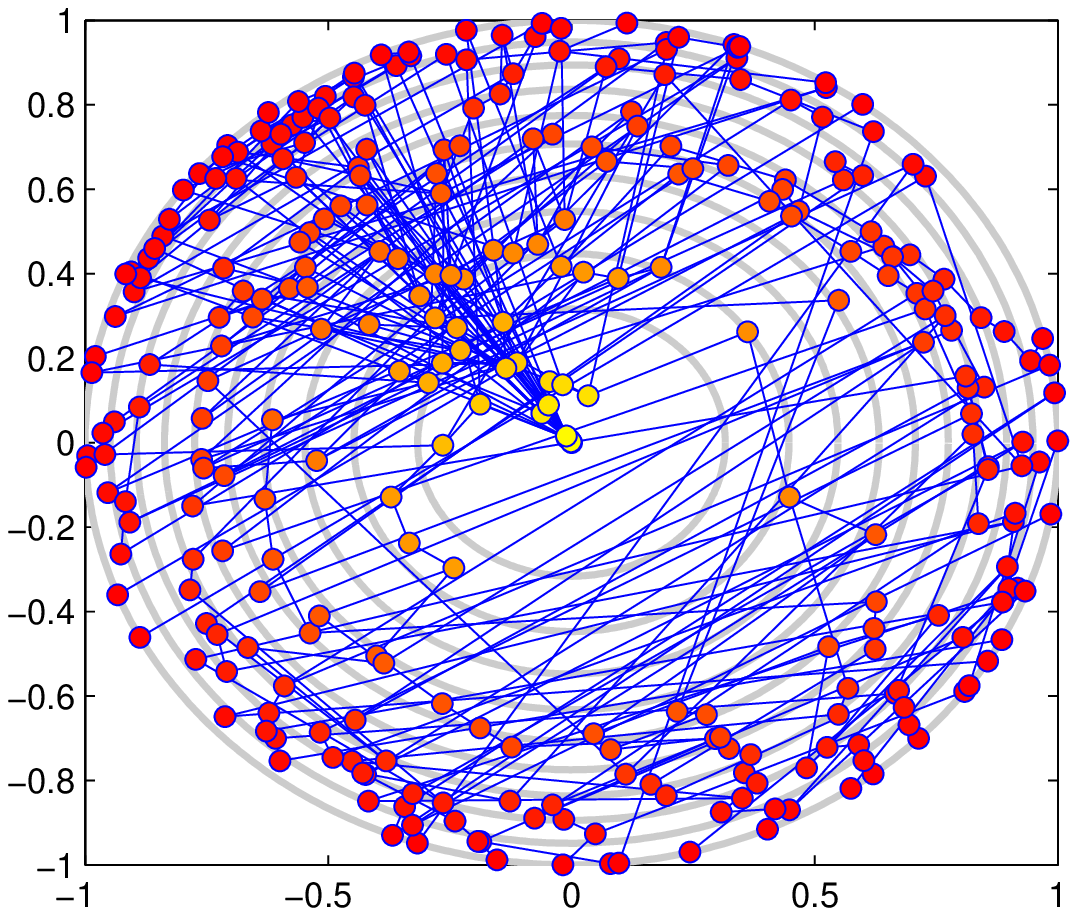}}
  \vspace*{-0.25cm}
  \centerline{(a) $\lambda = 0$}\medskip
  \vspace*{-0.25cm}
\end{minipage}
\hfill
\begin{minipage}[b]{.48\linewidth}
  \centering
  \centerline{\includegraphics[width=3.2cm]{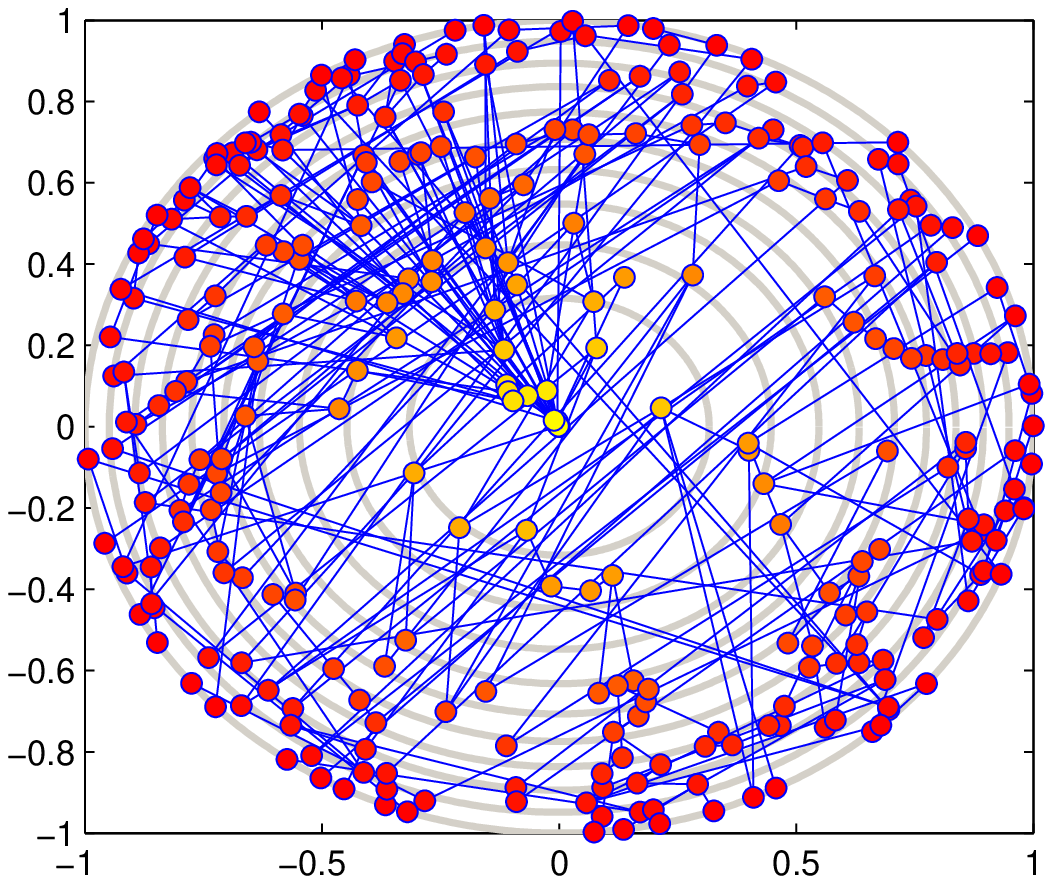}}
    \vspace*{-0.25cm}
  \centerline{(b) $\lambda = 1$}\medskip
  \vspace*{-0.25cm}
\end{minipage}
\begin{minipage}[b]{0.48\linewidth}
  \centering
  \centerline{\includegraphics[width=3.2cm]{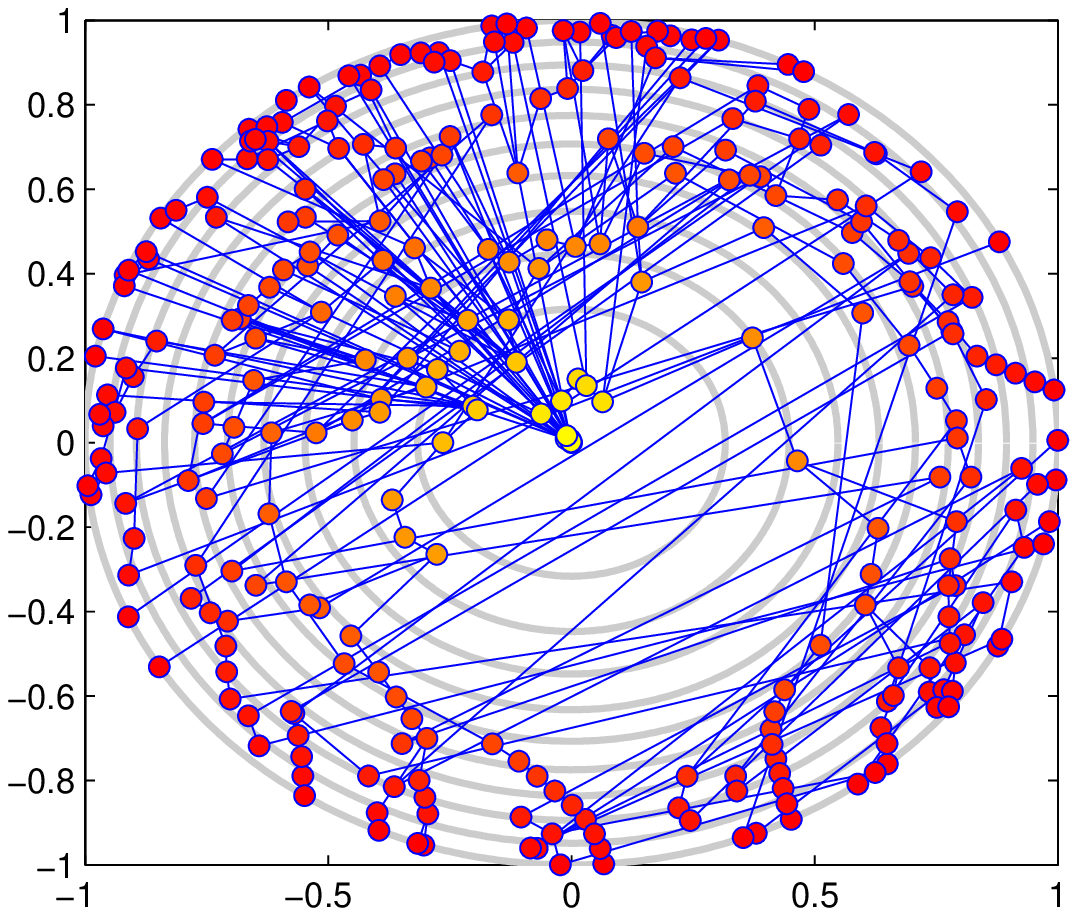}}
  \vspace*{-0.1cm}
  \centerline{(c) $\lambda = 100$}\medskip
\end{minipage}
\hfill
\begin{minipage}[b]{0.48\linewidth}
  \centering
  \centerline{\includegraphics[width=3.2cm]{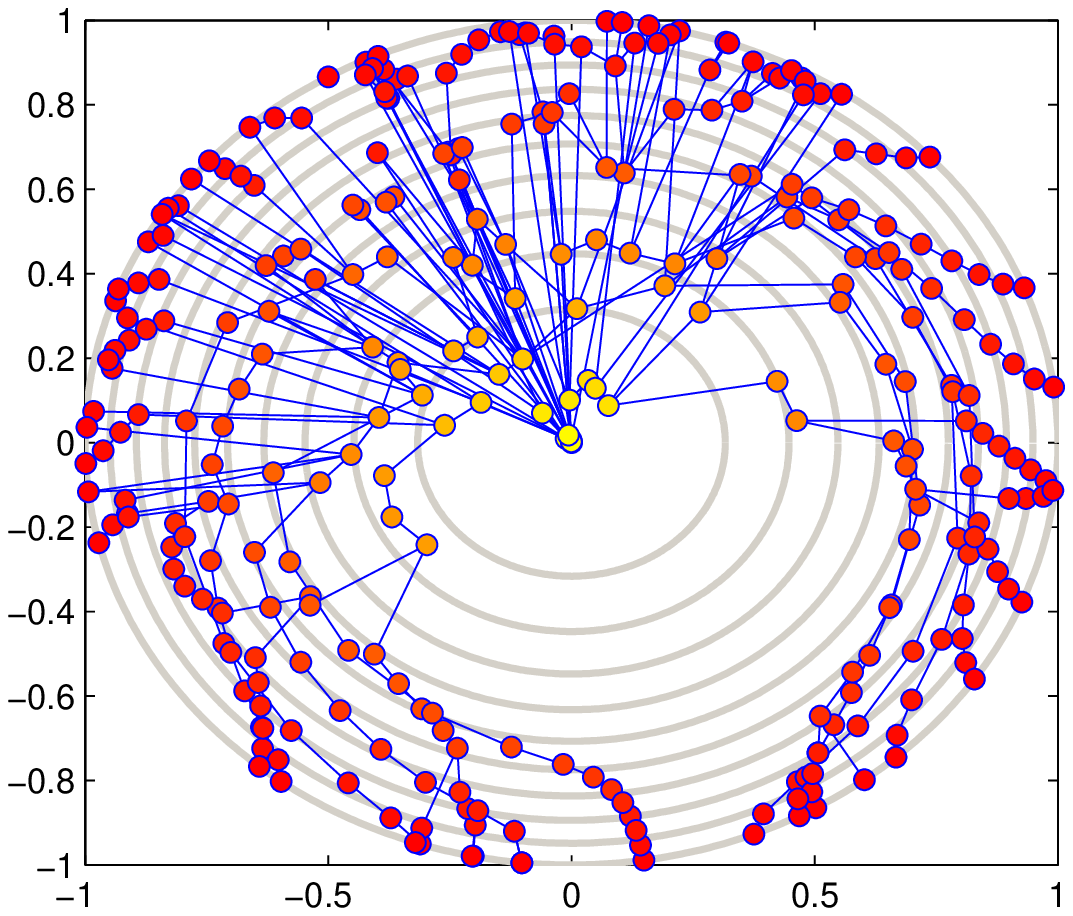}}
  \vspace*{-0.1cm}
  \centerline{(d) $\lambda = 10,000$}\medskip
\end{minipage}
\vspace*{-0.4cm}
\caption{Visualizing the London tube with a smoothing penalty}
\vspace*{-0.4cm}
\label{fig:varying_lambda}
\end{figure}

Figure \ref{fig:varying_lambda} illustrates the effect of including
the smoothness penalty. Increasing $\lambda$ promotes embeddings in
which edge crossings are minimized. This intuitively makes sense
because by forcing single-hop neighbors to lie close to each other,
the average edge length decreases, leading to fewer edge crossings.
In addition, increasing $\lambda$ yielded embeddings that were
aesthetically more appealing under fewer iterations.  For instance,
setting $\lambda = 10,000$ required only $30$ iterations for a
visualization that is comparable to running $150$ iterations with
$\lambda = 0$. An application of this work is travel time
cartography in which edge lengths reflect the amount of time it
takes to travel between stations. In this case, $f(c_i)$ is
equivalent to the transit time from a station of interest to any
other station $i$. By selecting a station and specifying the travel
time to all other nodes, an informative radial map centered at the
station of interest can be generated.
\begin{figure}[!htb]
\centering
\includegraphics[scale=0.275]{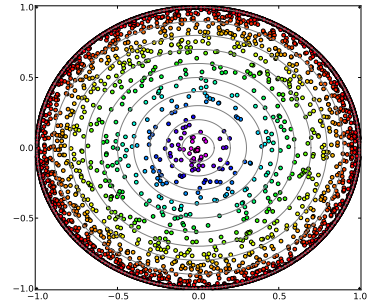}
\vspace*{-0.5cm} \caption{Embedding of a large social network}
\label{fig:arxiv_gen_relativity} \vspace*{-0.5cm}
\end{figure}

\subsection{Collaboration network of Arxiv General Relativity}
\label{arxiv} In this experiment, a large social network is
considered from the e-print arXiv repository covering scientific
collaborations between authors on papers submitted to the ``General
Relativity and Quantum Cosmology'' category (January
$1993$ to April $2003$) \cite{leskovec2007graph}. The nodes represent 
authors and an edge exists between nodes $i$ and $j$ if authors 
$i$ and $j$ co-authored a paper. Although the network contains $5,242$ nodes,
the embedding considered only its largest strongly connected
component comprising $4,158$ nodes. The objective was to embed the
network so that authors whose research is most related to the
majority of the others are placed closer to the center.

This behavior is best captured by the \emph{closeness centrality} that 
is defined as $c_i := (\sum_{j \in \mathcal{V}} d_{ij})^{-1}$, 
where $d_{ij}$ is the geodesic distance (lowest sum of edge weights) 
between nodes $i$ 
and $j$; and captures the extent to which any node lies close to all 
other nodes~\cite[p. 88]{netsci}. 
An informative mapping was obtained within $30$
outer iterations. For clarity and emphasis of the node positions,
edges were not included in the visualization. Drawings of graphs as
large as the autonomous systems within the Internet typically thin
out most of the edges.

Figure \ref{fig:arxiv_gen_relativity} shows the embedding with color
coding reflecting variations in centrality measure. The
proposed approach based on first-order methods leads to a fast
algorithm for visualizing such large networks. \vspace{-0.3cm}
\section{CONCLUSIONS}
\label{sec:conclusion} In this work, MDS-based means of embedding
graphs with certain structural constraints were proposed. In
particular, an optimization problem was formulated under centrality
constraints that are used to capture relative levels of importance
between the nodes. A block coordinate descent solver with successive
approximations was developed to deal with the non-convexity and
non-smoothness of the constrained MDS stress minimization problem.
In addition, a smoothness penalty term was incorporated to minimize
the edge crossings in the resultant network visualizations. Tests on
real-world networks were run and the results demonstrated that
convergence is guaranteed, and large networks can be visualized
relatively fast.

\end{document}